\documentclass[conference, 10 pt]{ieeeconf}  

\IEEEoverridecommandlockouts                              

\usepackage{subcaption}

\usepackage{graphicx} 
\usepackage{tikz}
\usepackage{pgfplots}
\usepackage{amsmath} 
\usepackage{amssymb}  
\usepackage{import}
\usepackage{dsfont}
\usepackage{hyperref}
\usepackage{gensymb}
\usepackage{booktabs}
\usepackage{tabularx}

\title{\LARGE \bf
Motion Estimation in Occupancy Grid Maps in Stationary Settings Using Recurrent Neural Networks
}

\author{Marcel Schreiber$^{1}$, Vasileios Belagiannis$^{1}$, Claudius Gl\"aser$^{2}$ and Klaus Dietmayer$^{1}$
	\thanks{The authors are with:}%
	\thanks{$^{1}$Institute of Measurement, Control, and Microtechnology, Ulm University, Germany, {\tt\footnotesize \{first.last\}@uni-ulm.de}}
	\thanks{$^{2}$Robert Bosch GmbH, Corporate Research, 71272 Renningen, Germany, {\tt\footnotesize \{first.last\}@de.bosch.com}}%
}
\begin{document}

\bstctlcite{IEEEexample:BSTcontrol} 

\def\cred{\textcolor{red}}
\def\cblue{\textcolor{blue}}
\def\cgreen{\textcolor{green}}

\newcommand\copyrighttextinitial{%

	\scriptsize This work has been submitted to the IEEE for possible publication. Copyright may be transferred without notice, after which this version may no longer be accessible.}%
\newcommand\copyrighttextfinal{%
	
	\scriptsize\copyright\ 2020 IEEE. Personal use of this material is permitted. Permission from IEEE must be obtained for all other uses, in any current or future media, including reprinting/republishing this material for advertising or promotional purposes,creating new collective works, for resale or redistribution to servers or lists, or reuse of any copyrighted component of this work in other works. DOI: 10.1109/ICRA40945.2020.9196702.}%
\newcommand\copyrightnotice{%

	\begin{tikzpicture}[remember picture,overlay]%

	\node[anchor=south,yshift=10pt] at (current page.south) {{\parbox{\dimexpr\textwidth-\fboxsep-\fboxrule\relax}{\copyrighttextfinal}}};%
	\end{tikzpicture}%


}

\maketitle
\copyrightnotice%
\thispagestyle{empty}
\pagestyle{empty}

\begin{abstract}
In this work, we tackle the problem of modeling the vehicle environment as dynamic occupancy grid map in complex urban scenarios using recurrent neural networks.
Dynamic occupancy grid maps represent the scene in a bird's eye view, where each grid cell contains the occupancy probability and the two dimensional velocity.
As input data, our approach relies on measurement grid maps, which contain occupancy probabilities, generated with lidar measurements.
Given this configuration, we propose a recurrent neural network architecture to predict a dynamic occupancy grid map, i.e. filtered occupancy and velocity of each cell, by using a sequence of measurement grid maps.
Our network architecture contains convolutional long-short term memories in order to sequentially process the input, makes use of spatial context, and captures motion. 
In the evaluation, we quantify improvements in estimating the velocity of braking and turning vehicles compared to the state-of-the-art.
Additionally, we demonstrate that our approach provides more consistent velocity estimates for dynamic objects, as well as, less erroneous velocity estimates in static area.
\end{abstract}
%
%
\section{Introduction}
Modeling and understanding the environment is essential for autonomous agents, such as self-driving vehicles.
A widely adopted environment representation is the occupancy grid map, where the vehicle's surrounding is divided into cells. Every cell is associated with a physical location and contains the occupancy probability, i.e. the probability to be occupied by an object.
In early works, grid maps had been mainly applied to robotic mapping due to the necessary assumption of a static environment~\cite{ElfesOccGrids}.

More recently, several works \cite{DanescuParticleBasedOccupancyGrid}, \cite{Negre_HybridSamplingBayesian}, \cite{TanzmeisterGridBasedMapping}, \cite{DBLP:journals/corr/NussRTYKMGD16} have extended the static grid map to dynamic environments, where the velocity has been introduced to the cell state.
The advantage of dynamic occupancy grid maps is the ability to represent arbitrary shaped objects, while there is no need for an explicit object detection and data association.
The aforementioned works use particle-based approaches, where particles in a cell represent the velocity distribution and the occupancy.
The particle-based dynamic occupancy grid maps have delivered good performance but they also come with some drawbacks.
The assumption of independent cells, in the update step of the particle filter, results in different velocity estimates in neighboring cells, even though they represent the same object.
Furthermore, a constant velocity model is usually assumed in the prediction step \cite{DanescuParticleBasedOccupancyGrid}, \cite{Negre_HybridSamplingBayesian}, \cite{TanzmeisterGridBasedMapping}, \cite{DBLP:journals/corr/NussRTYKMGD16}, which leads to decreased performance in dynamic driving scenarios, e.g.~brake and turn maneuvers. 

In this work, we present a learning-based approach for estimating the environment around the self-driving vehicle as a dynamic occupancy grid map.
A recurrent neural network is introduced for predicting the occupancy probability $P_O$ and the motion of the occupied cells, i.e.~the two velocity components directed to east and north $v_E, v_N$.  
\begin{figure}
	\centering
	\includegraphics[width=\columnwidth, trim=30 0 0 0, clip]{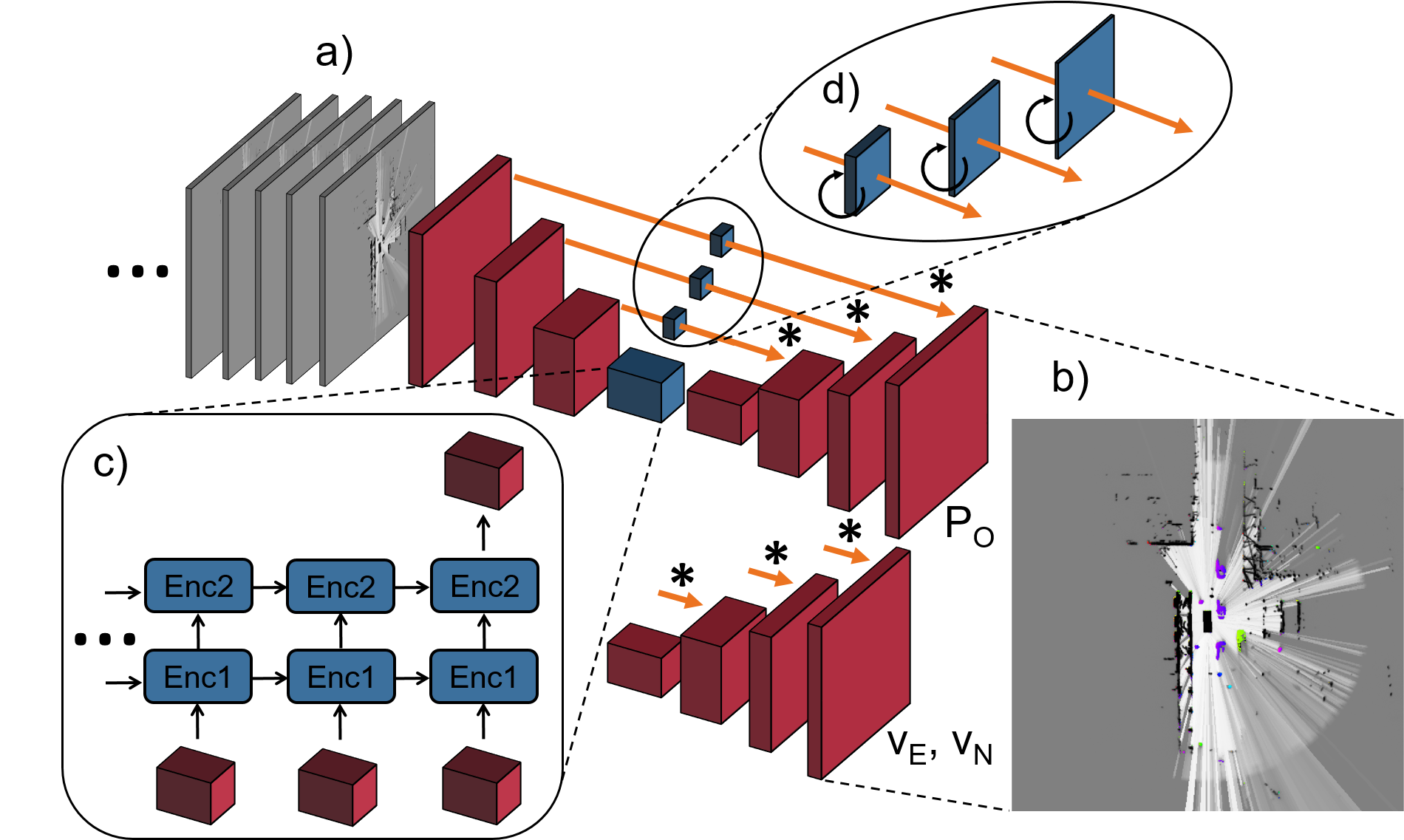}
	\caption{Illustration of our approach. Based on a sequence of measurement grid maps a) the network predicts occupancy and velocity, i.e. a dynamic occupancy grid map b). 
		Our network is a combination of feedforward (red) and recurrent (blue) network layers. 
		The Encoder-LSTM, outlined in c), uses information of a sequence of measurement grid maps to update the internal states and estimates occupancy and velocity.
		The recurrent skip architecture d) ensures dense predictions at the output.
		The two upscaling paths use the same skip connections, indicated by the stars.
	}
	\label{fig:teaser}
\end{figure}
The input data of our approach are occupancy grid maps based on lidar measurements of a single time step, hereinafter referred as measurement grid maps.
Our network architecture (see Fig. \ref{fig:teaser}) combines feedforward with recurrent modules to fuse information of past measurement grid maps with the current time step.
For the recurrent network part, convolutional long short-term memories (ConvLSTMs) \cite{DBLP:journals/corr/ShiCWYWW15} are used to capture spatio-temporal information in a sequence of measurement grid maps.
Unlike filtering approaches, the recurrent neural network (RNN) has the advantage of directly learning the motion from the data. It does not rely on motion assumptions such as the constant velocity model of the particle filter.
In addition, the utilization of convolution operations on the cells allows to use spatial context and consequently model their dependencies. 
Although the complete model is trained with supervision, we propose a pipeline for automatic label generation.
To that end, we collect large amount of laser measurements that are pre-processed and then labeled with the support of existing algorithms~\cite{DBLP:journals/corr/NussRTYKMGD16},~\cite{HoermannDynamicOccupancyGridPrediction},~\cite{Stumper18LabelExtraction}.

Our evaluation shows that our approach delivers rigid object motion estimates, as well as, robust velocity estimates in dynamic and static areas.
Finally, we compare our results quantitatively with the related work in three different driving scenarios to show that our learning-based approach is advantageous compared to filtering approaches.
In summary, our work makes the following contributions:
\begin{itemize}
	\item A combination of feedforward with recurrent neural network modules to predict dynamic occupancy grid maps.
	\item A data-driven approach that does not rely on motion assumptions.
	\item Promising performance compared to prior work on rigid motion prediction in dynamic driving scenarios such as vehicle brake and turn maneuvers.
\end{itemize}
\section{Related Work}
\noindent \textbf{Occupancy grid maps} are a common data representation in robotics for robust environment perception~\cite{ElfesOccGrids}, \cite{Thrun:2005:PR:1121596}. 
In earlier works, a binary Bayes filter has been usually employed to combine the measurements of subsequent time steps to a cell estimate, where the cell state is defined as occupied or free under the assumption of a static environment \cite{ElfesOccGrids}.
Occupancy grid maps have been also made popular in autonomous driving for environment representation, free space detection or collision avoidance \cite{Glaeser_EnvironmentPErceptionForInnerCityDriverAssistanceAndHighlyAutomatedDriving}.
However, the static environment assumption does not hold due to moving road users, such as pedestrians, vehicles and cyclists.
Several works \cite{DanescuParticleBasedOccupancyGrid}, \cite{Negre_HybridSamplingBayesian}, \cite{TanzmeisterGridBasedMapping}, \cite{DBLP:journals/corr/NussRTYKMGD16} have proposed to extend static occupancy grid maps to dynamic occupancy grid maps, mainly based on particle filter approaches.
Danescu \textit{et~al.}~\cite{DanescuParticleBasedOccupancyGrid} suggest to estimate position and velocity as continuous distribution using particles that can freely move between cells.
This idea has been further explored by Tanzmeister \textit{et~al.}~\cite{TanzmeisterGridBasedMapping} and N\`egre \textit{et~al.}~\cite{Negre_HybridSamplingBayesian} by representing only the dynamic environment with particles.
More recently, Nuss \textit{et~al.}~\cite{DBLP:journals/corr/NussRTYKMGD16} defined the dynamic state estimation as a random finite set problem and implemented particles in the Dempster-Shafer domain.
Unlike the prior work, we aim to estimate the occupancy and the velocity, based on recurrent neural networks.
The closest approach to our environment representation is \cite{DBLP:journals/corr/NussRTYKMGD16}, where we share the same measurement grid maps as input.
Thus, the algorithm of Nuss \textit{et~al.}\cite{DBLP:journals/corr/NussRTYKMGD16} is used as baseline to evaluate our approach and in our preprocessing algorithm to automatically generate labels.
\noindent \textbf{Deep learning} approaches, applied on grid maps have showed promising results due to the image-like data structure, which allows to rely on standard convolutional neural network architectures \cite{DiFengSurvey2019}.
Dynamic occupancy grid maps are utilized in~\cite{HoermannDetectionOnDynamicGridMaps} and~\cite{EngelDeepObjectTracking} to address the task of object detection; and in \cite{PiewakDynamicObjectDetection} to refine the separation of dynamic and static cells.
Wirges \textit{et~al.}~\cite{WirgesObjectDetectionAndClassification} propose a multi-layer grid to encode several features of lidar point clouds for object detection. 
The publications \cite{Song2DLidarMapPredictionsViaEstimatingMotionFLow} and \cite{WirgesSelfSupervisedFlowEstimation} propose a self-supervised flow estimation in grid maps to solve the task of future prediction, respectively odometry estimation.
The task of predicting future grid maps, as presented in \cite{HoermannDynamicOccupancyGridPrediction}, is closely related to the estimation of dynamics.
The grid prediction approaches in \cite{HoermannDynamicOccupancyGridPrediction} and \cite{SchreiberLongTermOccupancyGridPrediction} rely on dynamic occupancy grid maps input, which simplifies the prediction task by providing velocity estimates of each cell.
In contrast, we aim to estimate dynamic occupancy grid maps using only measurement grid maps as input in this work.
Recently, Hoyer \textit{et~al.}~\cite{HoyerShortTermPredictionSemanticGrids} proposed an architecture to generate a semantic grid representation using camera data and perform a short-term prediction. 

In \cite{DequaireDeepTrackingInTheWild} an end-to-end trainable framework is proposed to predict un-occluded occupancy grid maps in an unsupervised manner, based on recurrent neural networks.
The basic idea is to capture the dynamic evolution of the scene in a sequence of occupancy grid maps with a recurrent neural network. 
We differentiate our work by not only providing a filtered occupancy grid map, but also velocity estimates for the occupied cells.
Furthermore, the data in \cite{DequaireDeepTrackingInTheWild} is represented as grid maps with a size of $100 \times 100$, which is significantly smaller than our data representation, with a size of $901 \times 901$.
\section{Methods}\label{sec:methods}
In this section, we present our approach to motion estimation in occupancy grid maps.
We rely on a sequence of measurement grid maps as input to predict the dynamic occupancy grid map, i.e. occupancy probabilities and velocities.
To learn this mapping, we propose a recurrent neural network architecture, which we present in this section. 
In addition, we discuss the input and output data representation, the label generation and our objective function.
\subsection{Measurement Grid Maps}\label{sub:MeasGrids}
A grid map divides the vehicle environment into individual cells, where every cell contains information about the space located at its position.
In occupancy grid maps, every cell is described by the state $o_{k}$ that indicates whether the cell is free or occupied at the discrete time $k$~\cite{ElfesOccGrids}, \cite{Thrun:2005:PR:1121596}. 
We define the measurement grid map as the occupancy grid map based on the sensor measurements of a single time step.

In our setup, the occupancy probability of a cell in the measurement grid map is estimated from laser measurements $z_{k} = z_{1:n,k} = z_{1,k} \dots z_{n,k}$ with $n$ lidar beams of the single time step $k$, by applying the inverse sensor model \cite{Thrun:2005:PR:1121596}. 
The inverse sensor model assigns each grid cell $c$ an individual occupancy probability $p^{(c)}_{z_{i,k}}(o_{k}|z_{i,k})$, based on the laser measurement $z_{i,k}$.
Thus, grid cells $c$ near lidar beam reflections are detected as occupied with $p^{(c)}_{z_{i, k}}(o_{k}|z_{i,k})>0.5$, while the cells $c$ between the reflection and sensor are free space with $p^{(c)}_{z_{i,k}}(o_{k}|z_{i,k})<0.5$.
The occluded space outside the field of view are assigned with $p^{(c)}_{z_{i,k}}(o_{k}|z_{i,k})=0.5$.
The measurement grid map of time step $k$ combines in each cell $c$ information of $n$ individual laser beams using the binary Bayes filter
\begin{equation}\label{eq:Bayes}
p^{(c)}_{z_{k}}(o_{k}|z_{k})=\eta p^{(c)}_{z_{n,k}}(o_{k}|z_{n,k})p^{(c)}_{z_{1:n-1,k}}(o_{k}|z_{1:n-1,k}).
\end{equation}
Eq.~\ref{eq:Bayes} is used to recursively calculate the joint occupancy probability $p^{(c)}_{z_{k}}(o_{k}|z_{k})$, with the joint probability $p^{(c)}_{z_{1:n-1,k}}(o_{k}|z_{1:n-1,k})$ based on lidar beams up to $n-1$, the $n$-th measurement-based estimate $p^{(c)}_{z_{n,k}}(o_{k}|z_{n,k})$ and the normalization constant $\eta$ \cite{Thrun:2005:PR:1121596}.
The first processed lidar beam is used as initial prior
\begin{equation}\label{eq:InitialPosterior} 
	p^{(c)}_{z_{1:1,k}}(o_{k}|z_{1:1,k}) = p^{(c)}_{z_{1,k}}(o_{k}|z_{1,k}).
\end{equation} 
An example of a measurement grid map is depicted in Fig.~\ref{fig:trainingdata}.
The generation of measurement grid maps compared to working directly on lidar point clouds has several advantages. 
The data becomes structured, their dimensions are reduced and thus the processing by neural networks can be very efficient.
\subsection{Dynamic Occupancy Grid Maps}
The measurement grid map relies on the laser measurements of a single time step to represent the environment. 
In dynamic occupancy grid maps (DOGMs), the measurement grid map forms the input to the particle filter's update step.
Thus, information of several time steps are used to obtain a reliable estimation of occupied and free area; and the dynamics in the grid maps. 
An example of a DOGM is illustrated in Fig.~\ref{fig:trainingdata}.
The DOGM in \cite{DBLP:journals/corr/NussRTYKMGD16} employs a Dempster-Shafer \cite{Dempster2008} representation, thus the channels $\boldsymbol{C}=\{M_O, M_F, v_E, v_N, \sigma_{v_E}^2,\sigma_{v_N}^2, \sigma_{v_E,v_N}^2\}$ contain the mass for occupied $M_O \in [0,1]$ and the mass for free $M_F \in [0,1]$.
Furthermore, the velocities to the east $v_E$ and north $v_N$, as well as their variances and covariance are stored.
The corresponding occupancy probability $P_O$ is defined as $P_O = M_O + 0.5(1-M_O-M_F)$.
\subsection{Recurrent Network Architecture}
Our network architecture is a combination of feedforward and recurrent modules similar to the architecture in \cite{SchreiberLongTermOccupancyGridPrediction}, as depicted in Fig. \ref{fig:teaser}.
During training, it takes a sequence of measurement grid maps as input (Fig. \ref{fig:teaser} a) and predicts a dynamic occupancy grid map (Fig. \ref{fig:teaser} b), i.e. the occupancy $P\textsubscript{O}$ and the velocity components $v_E, v_N$ directed to the east and north for each cell of the current time step.
The size of the input data provided in $\mathds{R}^{901 \times 901 \times 4}$ is reduced to $\mathds{R}^{34 \times 34 \times 128}$ in three steps using convolutional layers with a stride of three, before feeding it in the LSTM-layer.
The Encoder-LSTM is depicted in Fig. \ref{fig:teaser} c) with the unrolled structure during training and consists of a two-layer ConvLSTM \cite{DBLP:journals/corr/ShiCWYWW15} with kernels of the size $3\times3$, and internal states $\boldsymbol{h,c}$ with the same size as the input.
During training, the Encoder-LSTM uses information of a sequence of measurement grid maps to sequentially update its internal states.
The output of the current time $k$ is then scaled up using several transposed convolutional layers to generate the network output layer with the channels $\boldsymbol{c}=\{P_{O}, v_E, v_N\}$ and the same spatial size as the measurement grid maps.
Here, separate upscaling layers are used for the occupancy $P_{O}$ and the velocities $v_E, v_N$.
So, the recurrent layers have to combine occupancy information of several single measurement grid maps to estimate a filtered occupancy probability, as produced by the particle filter approach.
Furthermore, the network detects groups of occupied cells and captures their motion in a sequence of measurement grid maps.
The estimation of two velocity components $v_E, v_N$ directed to the east and north is feasible, because the grid map is always aligned to these directions and so are the convolutional kernels.
Skip connections, as proposed in \cite{DBLP:journals/corr/LongSD14} and \cite{DBLP:journals/corr/RonnebergerFB15}, are also used to provide lower feature maps as additional input to the transposed convolutional layers to get dense predictions.
Instead of standard skip connections, we included ConvLSTMs in every skip connection, as depicted in Fig. \ref{fig:teaser} d).
This recurrent skip architecture provides a sequential filtering of high resolution features that is needed for modeling areas, that are temporary occluded in the measurement grid maps.
\subsection{Label Generation Process}\label{subsec:Training}
We propose to automatize the labeling process by employing a number of algorithms. 
We compute the ground truth DOGM with the algorithm of \cite{DBLP:journals/corr/NussRTYKMGD16} and additional data processing based on~\cite{HoermannDynamicOccupancyGridPrediction} and~\cite{Stumper18LabelExtraction}.

From~\cite{DBLP:journals/corr/NussRTYKMGD16}, we only make use of the occupancy probability $P_O$ of the DOGM, as occupancy label.
To obtain the velocity ground truth, we generate two binary masks for each time step, one for the static and one for the dynamic environment.
The classification of static cells is based on the algorithm described in \cite{HoermannDynamicOccupancyGridPrediction}, which relies on the occupancy probabilities of the DOGM.
Here, the main idea is to observe the occupancy probability of a single cell for a longer period of time and classify this cell, based on the variation of the occupancy value.
The static mask is then used to set the velocity of the cells in the static environment to zero.

Finally, we use the label algorithm of Stumper \textit{et~al.}\cite{Stumper18LabelExtraction} to automatically extract bounding boxes in grid map sequences.
This method firstly detects possible dynamic cells and generates object hypotheses, i.e. bounding boxes.
These boxes are then tracked forward and backward in time to verify the assumption of a dynamic object and to correct the shape, orientation and position of the box.
We use these bounding boxes to classify cells inside bounding boxes, that contain an occupancy probability $P_O > 0.55$ as dynamic cells.
We further use the displacement of these bounding boxes to get the values of our velocity label $v_E, v_N$ for the dynamic cells.
Our generated velocity labels have uniform velocity values per object. 
This helps to learn predicting rigid motion in contrast to the particle filter estimates from~\cite{DBLP:journals/corr/NussRTYKMGD16}.
We show this in the evaluation part by comparing the velocity estimates of our approach with the estimates of the particle filter.

To sum up, the velocity estimates are obtained in two steps: firstly, the velocity values are set to zero in the static environment; secondly, the velocity label of dynamic cells is corrected to a uniform motion per object.
Note that we make use of cells for updating the neural network parameters only if they are certainly classified as either static or dynamic, as further described in the following section.
\subsection{Loss Function}\label{subsec:Loss}
The prediction of occupancy and both velocity components are the input to our loss functions. 
The occupancy loss $L\textsubscript{P\textsubscript{O}}$ is defined as:
\begin{equation}
L\textsubscript{P\textsubscript{O}} = \frac{1}{W\cdot H} \sum_{i=1}^{W\cdot H} \lambda\textsubscript{P\textsubscript{O},c}(i)L\textsubscript{P\textsubscript{O},c}(i).
\end{equation}
where $W$ is the width and $H$ the height of the processed grid map. The actual loss per cell L\textsubscript{P\textsubscript{O},c}(i) is the Huber loss, given by
\begin{equation}
\begin{split}
&L\textsubscript{P\textsubscript{O},c}(i) = \\ 
	&\left\{
		\begin{aligned}
		0.5(y\textsubscript{P\textsubscript{O}}(i)-\hat{y}\textsubscript{P\textsubscript{O}}(i))^2 \hspace{0.8em} &\text{if}\ |y\textsubscript{P\textsubscript{O}}(i)-\hat{y}\textsubscript{P\textsubscript{O}}(i)| \leq \delta \\
		\delta|y\textsubscript{P\textsubscript{O}}(i)-\hat{y}\textsubscript{P\textsubscript{O}}(i)|-0.5\delta^2 \hspace{0.8em} &\text{otherwise}, 										
		\end{aligned}
	\right.
\end{split}
\end{equation}
where $y\textsubscript{P\textsubscript{O}}(i)$ is the occupancy label and $\hat{y}\textsubscript{P\textsubscript{O}}(i)$ the  occupancy prediction  of cell $i$. We empirically set the hyperparameter $\delta$ to $0.02$.
Finally, the cell-wise weighting term $\lambda\textsubscript{P\textsubscript{O},c}(i)$ is defined as 
\begin{equation}
	\lambda\textsubscript{P\textsubscript{O},c}(i) =\left\{
		\begin{aligned}
			y\textsubscript{P\textsubscript{O}}(i)\lambda\textsubscript{p} \quad &\text{if}\ y\textsubscript{P\textsubscript{O}}(i) > 0.5 \\
			(1-y\textsubscript{P\textsubscript{O}}(i))\lambda\textsubscript{p} \quad &\text{if}\ y\textsubscript{P\textsubscript{O}}(i) < 0.5 \\
			1 \quad &\text{if}\ y\textsubscript{P\textsubscript{O}}(i) = 0.5,
		\end{aligned} 
	\right.	
\end{equation}
where we also empirically set $\lambda\textsubscript{p}$ to $4$. The weighting term ensures that the occupancy labels closer to 0 or 1 have a higher impact on the optimization. 

Next, the velocity loss $L\textsubscript{v}$ is defined as
\begin{equation}\label{eq:velocityLoss}
L\textsubscript{v} = \frac{1}{W \cdot H} \sum_{i=1}^{W \cdot H} 0.5\lambda\textsubscript{v,c}(i)(y\textsubscript{v}(i)-\hat{y}\textsubscript{v}(i))^2,
\end{equation}
where $y\textsubscript{v}(i)$ corresponds to the velocity label and $\hat{y}\textsubscript{v}(i)$ to the velocity prediction of cell $i$.
%
%
The weighting factor $\lambda\textsubscript{v,c}(i)$ is set to the value $\lambda\textsubscript{s} = 5$ for cells belonging to the static mask and $\lambda\textsubscript{d} = 20$ for cells in the dynamic mask.
For occupied cells, which are neither dynamic or static, as well as, for free or unobserved cells, the weighting term $\lambda\textsubscript{v,c}(i)$ is set to zero and thus these areas do not contribute to the parameter update of our model.
Therefore, the network is not explicitly trained to predict zero velocity in free space, since we can easily correct this by considering only velocity estimates for occupied cells.
During deployment, we generate a binary mask using the occupancy prediction with a threshold $P\textsubscript{O} > 0.55$ and perform an element-wise multiplication of this mask with the velocity predictions.
The velocity loss $L\textsubscript{v}$ is used for the losses of both velocity components with the same weighting masks.

The complete loss function is a sum of the occupancy loss $L\textsubscript{P\textsubscript{O}}$ and the velocity loss in east $L\textsubscript{v,E}$ and the velocity loss in north $L\textsubscript{v,N}$
\begin{equation}\label{eq:overallLoss}
L\textsubscript{o} = \alpha\textsubscript{p}L\textsubscript{P\textsubscript{O}} + \alpha\textsubscript{v}(L\textsubscript{v,E}+L\textsubscript{v,N}),
\end{equation}
where the weight factors $\alpha\textsubscript{p}$ and $\alpha\textsubscript{v}$ determine the balance between the loss terms, i.e. occupancy and velocity prediction.
We choose factors of $\alpha\textsubscript{p} = 10, \alpha\textsubscript{v} = 0.5$ to encounter the small values of the Huber-Loss with $\delta = 0.02$ in the occupancy loss.
We train our neural network based on the loss term specified in Eq.~\ref{eq:overallLoss} with back-propagation and stochastic gradient descent.
\section{Experimental Setup}
Our work is focused on stationary settings where the vehicle has zero ego-motion. To examine this scenario, we rely on our own dataset due to the lack of standard public benchmarks. Next, we present our dataset and also the implementation details of our model.
\subsection{Dataset}
\begin{figure}
	\centering
	\vspace{1.42mm}
	\begin{subfigure}{0.49\columnwidth}
		\includegraphics[width=\columnwidth]{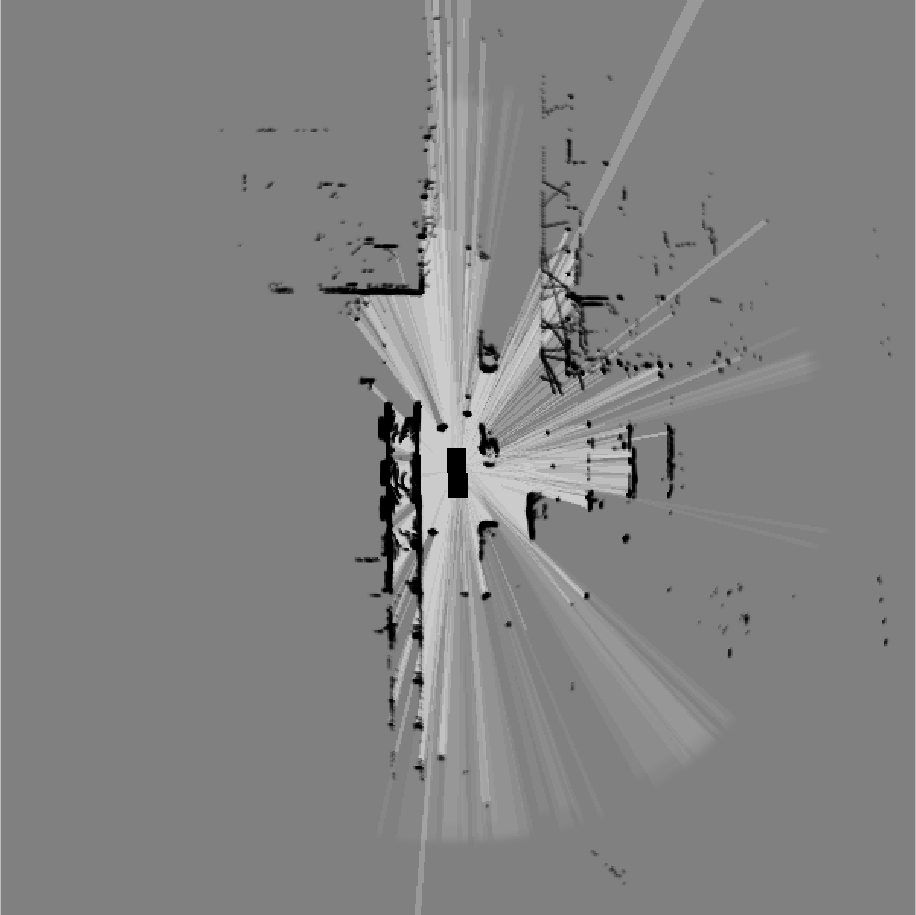}
		\label{fig:trainingdataMeas}
	\end{subfigure}
	\begin{subfigure}{0.49\columnwidth}
		\includegraphics[width=\columnwidth]{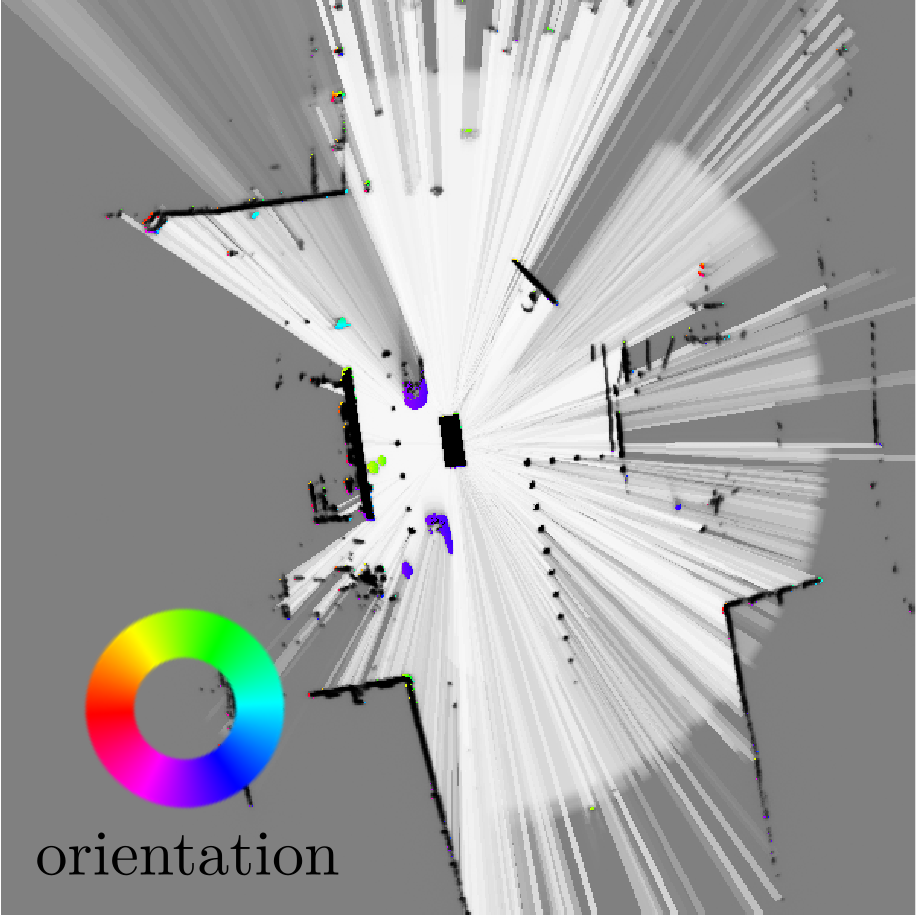}
		\label{fig:trainingdataDOGM}
	\end{subfigure}
	\caption{Illustration of the two intersections used for training. Left: measurement grid map, right: dynamic occupancy grid map. 
	The occupancy probability $P_O$ is encoded in gray-scale, where occupied cells are depicted in \textit{black}, free space is \textit{white} and the unobserved area is \textit{gray}.
	Note that our ego-vehicle is the black rectangle in the center.	In the dynamic occupancy grid map, moving cells are illustrated with colors, according to the orientation of the velocity.}
	\label{fig:trainingdata}
\end{figure}
We recorded training data of about 2\,h of two busy urban intersections, depicted in Fig. \ref{fig:trainingdata}, with the presence of pedestrians, cyclists and vehicles, from a stationary vehicle.
Our experimental vehicle was equipped with an IBEO LUX long-range lidar scanner in the front of the vehicle and four spinning Velodyne VLP-16 laser scanners, providing $360\degree$ perception with a frequency of 10\,Hz.
Based on the lidar measurements, we generated for every time step two measurement grid maps, one for each sensor type, and the dynamic occupancy grid map.
The dynamic occupancy grid map is used to generate our training labels as described in Section \ref{subsec:Training}.
The input and output resolution of the grid maps cover $901\times901$ cells where a cell has 0.15\,m width, leading to a total perception area of $135.15\,\text{m}\times135.15\,\text{m}$.
In addition, we made recordings of three different urban intersections, one for validation and two for testing.
The scenes of the two test sequences are depicted in Fig. \ref{fig:qualitativeEval} and are used to examine the generalization capability of our approach and a qualitative evaluation.
Since we only obtain automatically generated labels in these test scenarios, we have additionally recorded three sequences in which a second experimental vehicle equipped with a differential global positioning system (DGPS) provides precise ground truth motion data.
With these sequences, we evaluate the performance in different driving situations: driving straight with constant velocity, stop and go, and driving circles.
Due to the applied constant velocity model, the particle-based approach has advantages in the scenario with constant velocity.
However, we also evaluate the performance in more dynamic situations, since these situations occur in the application of our system, i.e. dynamic environment representation on urban intersections.
\subsection{Implementation}
For network parameter learning, we rely on the ADAM \cite{DBLP:journals/corr/KingmaB14} solver with exponential decay rates $\beta_1 = 0.9$ and $\beta_2 = 0.999$, with a learning rate $lr = 1e^{-4}$.
In addition, we normalize the velocity values $v\textsubscript{E}$, $v\textsubscript{N}$ with a factor of 15, to squash the velocity label in the same range as our occupancy labels.
We choose this factor based on the maximum expected velocity in our urban setting.
Our recurrent network is trained with sequences of $n\textsubscript{in}=10$ measurement grid maps. 
We use data augmentation, i.e. we rotate the grid maps randomly in $1\degree$ steps, which gives us the opportunity to cover all possible motion directions in our training data. 
With this augmentation, the use of only two intersections for training turned out to be sufficient to gain good performance in various environments.
In addition to data augmentation, we apply dropout on the non-recurrent connections of the Encoder-LSTM as proposed in \cite{DBLP:journals/corr/ZarembaSV14} to avoid overfitting.
The developed recurrent network architecture contains 3.7 million parameters and achieves an inference time of about 53\,ms on a Nvidia GeForce GTX 1080 Ti, which is sufficient for \mbox{real-time} application.
\section{Evaluation}
In our experiments, we consider the three test driving sequences: driving straight with constant velocity, stop and go, and driving circles.
In the quantitative evaluations, we calculate the mean velocity magnitude and the mean orientation of the cells belonging to one object, to compare these estimates with the ground truth data of the reference vehicle. 
In addition to the object velocity and orientation, we  compute the standard deviations of the velocity magnitude and orientation.
These standard deviations serve as a measure of the extent to which the estimates of the cells belonging to one object differ.
To give a summary of the performance on a whole sequence, we calculated mean values of the absolute errors in velocity magnitude MAE\textsubscript{vel} and orientation MAE\textsubscript{ori}. 
Finally, we take the average of the standard deviations for velocity magnitude $\bar{\sigma}\textsubscript{vel}$ and orientation $\bar{\sigma}\textsubscript{ori}$.
We compare our approach to a state-of-the-art particle filter approach from literature \cite{DBLP:journals/corr/NussRTYKMGD16}.
In the evaluation, we denote the results of the algorithm in \cite{DBLP:journals/corr/NussRTYKMGD16} as \emph{Nuss \textit{et~al.}}, our approach is stated as \emph{ours} and the ground truth as \emph{reference}.
After the quantitative evaluation, the main differences between the estimates of the particle filter approach and our work are qualitatively discussed in Section \ref{subsec:qualitativeEval}.
\subsection{Driving Straight with Constant Velocity}
\begin{table}
	\centering
	\vspace{1mm}
	\setlength{\tabcolsep}{0.17cm}
	\caption{Evaluation Summary}
	\label{tab:EvalSummary}
	\begin{tabular}{@{}ccccccc@{}}
		\toprule
		approach&scenario& duration& MAE\textsubscript{vel}&MAE\textsubscript{ori} &$\bar{\sigma}\textsubscript{vel}$&$\bar{\sigma}\textsubscript{ori}$  \\ 
		  &  & [s] & [m/s] & [\degree] & [m/s] & [\degree] \\
		\midrule
		ours & straight & 11 & 0.746 & 3.635 & \textbf{0.629} & \textbf{4.282} \\
		Nuss \textit{et~al.} & straight & 11 & \textbf{0.661} & \textbf{2.885} & 1.005 & 13.682\\
		\midrule
		ours & stop \& go & 15 & \textbf{0.742} & \textbf{8.447} & \textbf{0.264} & \textbf{8.248}  \\
		Nuss \textit{et~al.}  & stop \& go & 15 & 1.268 & 10.515 & 0.683 & 23.230 \\
		\midrule
		ours & circles & 24 & \textbf{0.687} & \textbf{6.205} & \textbf{0.554} & \textbf{9.006} \\
		Nuss \textit{et~al.}  & circles & 24 & 1.850 & 21.084 & 1.286 & 16.645 \\	
		\bottomrule
	\end{tabular}
\end{table}
First, we compare the results in a scenario where a test vehicle drives straight with a nearly constant velocity of approximately 5\,m/s.
The mean metrics in Table~\ref{tab:EvalSummary} show that the particle filter approach yields slightly better predictions for the velocity magnitude and orientation in this scenario.
Although, our approach provides lower mean standard deviations, i.e. the estimates are more uniform.
This result supports the observations in the qualitative evaluations, depicted in Fig.~\ref{fig:qualitativeEval}.
The enhanced performance of the particle filter approach can be explained with the setting of this scenario:
the vehicle drives with constant speed straight ahead and therefore exactly matches the constant velocity motion model that is applied in \cite{DBLP:journals/corr/NussRTYKMGD16}. 
\subsection{Stop and Go}
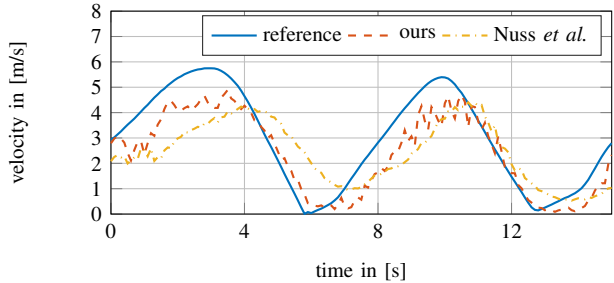
\begin{figure}
	\centering
	\newlength\fheight
	\newlength\fwidth
	\setlength\fheight{2.7cm}
	\setlength\fwidth{7cm}
%
%
\definecolor{mycolor1}{rgb}{0.00000,0.44700,0.74100}%
\definecolor{mycolor2}{rgb}{0.85000,0.32500,0.09800}%
\definecolor{mycolor3}{rgb}{0.92900,0.69400,0.12500}%
\definecolor{mycolor4}{rgb}{0.49400,0.18400,0.55600}%
%
%
\begin{tikzpicture}

\begin{axis}[%
width=0.951\fwidth,
height=\fheight,
at={(0\fwidth,0\fheight)},
scale only axis,
xmin=0,
xmax=15,
xlabel style={font=\color{white!15!black}},
xlabel={time in [s]},
ymin=0,
ymax=8,
xtick = {0, 4, 8, 12, 16},
xticklabels = {0, 4, 8, 12, 16},
ytick = {0, 1, 2, 3, 4, 5, 6, 7, 8},
yticklabels = {0, 1, 2, 3, 4, 5, 6, 7, 8},
xmajorgrids,
ymajorgrids,
ylabel style={font=\color{white!15!black}},
ylabel={velocity in [m/s]},
axis background/.style={fill=white},
legend style={legend cell align=left, align=left, draw=white!15!black},
legend style={font=\footnotesize, legend columns=3}, xticklabel style = {font=\footnotesize}, yticklabel style = {font=\footnotesize}, 
xlabel style = {font=\footnotesize}, ylabel style = {font=\footnotesize}
]
\addplot [color=mycolor1, thick]
  table[row sep=crcr]{%
0.000	2.932\\
0.100	3.023\\
0.200	3.141\\
0.300	3.259\\
0.400	3.395\\
0.500	3.524\\
0.600	3.642\\
0.699	3.761\\
0.800	3.901\\
0.900	4.039\\
0.999	4.180\\
1.100	4.306\\
1.200	4.452\\
1.300	4.592\\
1.400	4.714\\
1.500	4.846\\
1.600	4.943\\
1.700	5.070\\
1.800	5.172\\
1.900	5.275\\
2.000	5.363\\
2.100	5.430\\
2.200	5.511\\
2.300	5.567\\
2.400	5.621\\
2.500	5.664\\
2.601	5.700\\
2.700	5.716\\
2.800	5.745\\
2.901	5.747\\
3.000	5.747\\
3.100	5.741\\
3.200	5.708\\
3.300	5.647\\
3.400	5.567\\
3.500	5.466\\
3.600	5.346\\
3.700	5.205\\
3.800	5.055\\
3.900	4.865\\
4.000	4.649\\
4.100	4.432\\
4.200	4.195\\
4.300	3.949\\
4.400	3.684\\
4.500	3.415\\
4.600	3.139\\
4.700	2.884\\
4.800	2.641\\
4.900	2.380\\
5.000	2.119\\
5.100	1.854\\
5.200	1.580\\
5.300	1.318\\
5.400	1.061\\
5.500	0.803\\
5.600	0.538\\
5.700	0.264\\
5.800	0.006\\
5.899	0.079\\
6.000	0.038\\
6.100	0.104\\
6.199	0.162\\
6.300	0.214\\
6.400	0.282\\
6.500	0.354\\
6.600	0.431\\
6.700	0.538\\
6.800	0.659\\
6.900	0.833\\
7.000	1.005\\
7.100	1.204\\
7.200	1.399\\
7.300	1.581\\
7.400	1.771\\
7.500	1.964\\
7.600	2.144\\
7.701	2.306\\
7.801	2.481\\
7.901	2.649\\
8.001	2.820\\
8.101	2.979\\
8.201	3.148\\
8.301	3.319\\
8.401	3.498\\
8.501	3.669\\
8.601	3.847\\
8.701	4.013\\
8.801	4.173\\
8.902	4.323\\
9.002	4.480\\
9.102	4.627\\
9.202	4.773\\
9.302	4.905\\
9.402	5.038\\
9.502	5.154\\
9.602	5.249\\
9.702	5.318\\
9.802	5.369\\
9.902	5.397\\
10.002	5.387\\
10.102	5.343\\
10.202	5.243\\
10.302	5.114\\
10.402	4.935\\
10.502	4.779\\
10.602	4.572\\
10.702	4.370\\
10.802	4.141\\
10.902	3.917\\
11.002	3.679\\
11.102	3.459\\
11.202	3.229\\
11.302	2.995\\
11.401	2.776\\
11.501	2.549\\
11.601	2.337\\
11.701	2.114\\
11.801	1.895\\
11.901	1.681\\
12.001	1.469\\
12.101	1.270\\
12.201	1.066\\
12.300	0.868\\
12.400	0.652\\
12.500	0.448\\
12.600	0.266\\
12.700	0.165\\
12.800	0.145\\
12.900	0.199\\
13.000	0.255\\
13.100	0.316\\
13.200	0.375\\
13.300	0.442\\
13.399	0.502\\
13.499	0.570\\
13.599	0.634\\
13.699	0.682\\
13.798	0.754\\
13.899	0.821\\
13.998	0.913\\
14.098	1.019\\
14.198	1.174\\
14.298	1.350\\
14.398	1.575\\
14.498	1.797\\
14.598	2.029\\
14.698	2.255\\
14.798	2.455\\
14.898	2.641\\
14.998	2.794\\
};
\addlegendentry{reference}

\addplot [color=mycolor2, dashed, thick]
  table[row sep=crcr]{%
0.000	2.789\\
0.100	2.977\\
0.200	3.068\\
0.300	2.975\\
0.400	2.818\\
0.500	2.413\\
0.600	2.112\\
0.699	2.520\\
0.800	2.853\\
0.900	2.877\\
0.999	2.359\\
1.100	2.789\\
1.200	3.187\\
1.300	3.085\\
1.400	3.522\\
1.500	3.733\\
1.600	4.137\\
1.700	4.415\\
1.800	4.462\\
1.900	4.295\\
2.000	4.083\\
2.100	4.072\\
2.200	4.176\\
2.300	4.245\\
2.400	4.089\\
2.500	4.198\\
2.601	4.333\\
2.700	4.534\\
2.800	4.667\\
2.901	4.399\\
3.000	4.235\\
3.100	4.185\\
3.200	4.186\\
3.300	4.583\\
3.400	4.732\\
3.500	4.898\\
3.600	4.560\\
3.700	4.464\\
3.800	4.278\\
3.900	4.139\\
4.000	4.271\\
4.100	4.129\\
4.200	3.989\\
4.300	3.696\\
4.400	3.658\\
4.500	3.439\\
4.600	3.293\\
4.700	2.998\\
4.800	3.108\\
4.900	2.781\\
5.000	2.470\\
5.100	2.145\\
5.200	2.201\\
5.300	1.876\\
5.400	1.706\\
5.500	1.337\\
5.600	1.353\\
5.700	0.988\\
5.800	0.775\\
5.899	0.430\\
6.000	0.413\\
6.100	0.331\\
6.199	0.305\\
6.300	0.342\\
6.400	0.259\\
6.500	0.305\\
6.600	0.338\\
6.700	0.198\\
6.800	0.376\\
6.900	0.639\\
7.000	0.298\\
7.100	0.225\\
7.200	0.271\\
7.300	0.532\\
7.400	0.655\\
7.500	0.568\\
7.600	1.121\\
7.701	1.042\\
7.801	1.326\\
7.901	1.579\\
8.001	1.639\\
8.101	1.814\\
8.201	1.699\\
8.301	1.882\\
8.401	2.325\\
8.501	2.480\\
8.601	2.471\\
8.701	2.808\\
8.801	3.303\\
8.902	2.884\\
9.002	3.064\\
9.102	2.700\\
9.202	3.910\\
9.302	4.251\\
9.402	3.669\\
9.502	3.334\\
9.602	4.219\\
9.702	3.997\\
9.802	3.687\\
9.902	3.460\\
10.002	4.367\\
10.102	4.648\\
10.202	4.063\\
10.302	3.769\\
10.402	4.458\\
10.502	4.695\\
10.602	4.422\\
10.702	3.434\\
10.802	4.257\\
10.902	4.218\\
11.002	3.651\\
11.102	3.330\\
11.202	3.601\\
11.302	3.070\\
11.401	2.751\\
11.501	2.402\\
11.601	3.065\\
11.701	2.469\\
11.801	1.846\\
11.901	1.729\\
12.001	1.614\\
12.101	1.609\\
12.201	1.147\\
12.300	1.016\\
12.400	0.770\\
12.500	0.553\\
12.600	0.451\\
12.700	0.369\\
12.800	0.254\\
12.900	0.236\\
13.000	0.152\\
13.100	0.194\\
13.200	0.147\\
13.300	0.089\\
13.399	0.144\\
13.499	0.178\\
13.599	0.234\\
13.699	0.256\\
13.798	0.154\\
13.899	0.112\\
13.998	0.183\\
14.098	0.239\\
14.198	0.318\\
14.298	0.533\\
14.398	0.642\\
14.498	0.925\\
14.598	1.126\\
14.698	1.213\\
14.798	1.215\\
14.898	1.965\\
14.998	2.327\\
};
\addlegendentry{ours}

\addplot [color=mycolor3, dashdotted, thick]
  table[row sep=crcr]{%
0.000	2.085\\
0.100	2.252\\
0.200	2.324\\
0.300	2.234\\
0.400	2.285\\
0.500	1.994\\
0.600	2.111\\
0.699	2.234\\
0.800	2.163\\
0.900	2.596\\
0.999	2.521\\
1.100	2.275\\
1.200	2.587\\
1.300	2.095\\
1.400	2.243\\
1.500	2.191\\
1.600	2.506\\
1.700	2.626\\
1.800	2.602\\
1.900	2.863\\
2.000	2.982\\
2.100	3.046\\
2.200	3.236\\
2.300	3.215\\
2.400	3.308\\
2.500	3.365\\
2.601	3.440\\
2.700	3.411\\
2.800	3.570\\
2.901	3.561\\
3.000	3.689\\
3.100	3.785\\
3.200	3.616\\
3.300	3.815\\
3.400	3.839\\
3.500	4.051\\
3.600	3.968\\
3.700	4.046\\
3.800	4.145\\
3.900	4.236\\
4.000	4.158\\
4.100	4.165\\
4.200	4.095\\
4.300	4.140\\
4.400	4.026\\
4.500	4.067\\
4.600	4.046\\
4.700	3.990\\
4.800	3.840\\
4.900	3.808\\
5.000	3.629\\
5.100	3.555\\
5.200	3.212\\
5.300	3.104\\
5.400	3.079\\
5.500	2.860\\
5.600	2.661\\
5.700	2.556\\
5.800	2.429\\
5.899	2.210\\
6.000	2.007\\
6.100	1.807\\
6.199	1.705\\
6.300	1.628\\
6.400	1.415\\
6.500	1.324\\
6.600	1.245\\
6.700	1.151\\
6.800	1.081\\
6.900	1.100\\
7.000	1.023\\
7.100	0.947\\
7.200	0.974\\
7.300	1.020\\
7.400	1.011\\
7.500	1.060\\
7.600	1.171\\
7.701	1.267\\
7.801	1.399\\
7.901	1.523\\
8.001	1.516\\
8.101	1.591\\
8.201	1.674\\
8.301	1.774\\
8.401	1.801\\
8.501	1.882\\
8.601	1.886\\
8.701	1.938\\
8.801	2.045\\
8.902	2.149\\
9.002	2.381\\
9.102	2.435\\
9.202	2.568\\
9.302	2.803\\
9.402	2.811\\
9.502	2.941\\
9.602	3.077\\
9.702	3.367\\
9.802	3.626\\
9.902	3.740\\
10.002	3.758\\
10.102	4.043\\
10.202	4.035\\
10.302	4.218\\
10.402	4.093\\
10.502	4.230\\
10.602	4.233\\
10.702	4.404\\
10.802	4.257\\
10.902	4.388\\
11.002	4.263\\
11.102	4.301\\
11.202	3.847\\
11.302	3.608\\
11.401	3.264\\
11.501	3.488\\
11.601	2.817\\
11.701	2.642\\
11.801	2.582\\
11.901	2.437\\
12.001	1.963\\
12.101	1.719\\
12.201	1.808\\
12.300	1.747\\
12.400	1.420\\
12.500	1.318\\
12.600	1.212\\
12.700	1.124\\
12.800	0.946\\
12.900	0.904\\
13.000	0.846\\
13.100	0.772\\
13.200	0.691\\
13.300	0.665\\
13.399	0.608\\
13.499	0.568\\
13.599	0.569\\
13.699	0.587\\
13.798	0.531\\
13.899	0.581\\
13.998	0.624\\
14.098	0.596\\
14.198	0.582\\
14.298	0.664\\
14.398	0.636\\
14.498	0.702\\
14.598	0.750\\
14.698	0.797\\
14.798	0.924\\
14.898	1.012\\
14.998	1.034\\
};
\addlegendentry{Nuss \textit{et~al.}}
\end{axis}

\begin{axis}[%
width=1.227\fwidth,
height=1.227\fheight,
at={(-0.16\fwidth,-0.135\fheight)},
scale only axis,
xmin=0.000,
xmax=1.000,
ymin=0.000,
ymax=1.000,
axis line style={draw=none},
ticks=none,
axis x line*=bottom,
axis y line*=left,
legend style={legend cell align=left, align=left, draw=white!15!black},
legend style={font=\footnotesize}
]
\end{axis}
\end{tikzpicture}%
	\caption[Stop and go velocity magnitude]{Velocity magnitudes in the stop and go scenario.}
	\label{fig:StopAndGoVelMag}
\end{figure}
In this scenario, a reference vehicle accelerates, and stops several times, as illustrated in Fig.~\ref{fig:StopAndGoVelMag} (blue line). 
The results show, that our approach (red line) has the ability to capture braking and accelerating more accurate and with less delay than the particle-based algorithm.
However, the prediction accuracy of the orientation decreases for both approaches, compared to the previous scenario (see Table~\ref{tab:EvalSummary}). 
We explain this behavior with the output representation, where the two velocity components $v_E, v_N$ are estimated.
Thus, in a scenario with changing speed, there occur errors in the orientation estimates, even when the vehicle drives straight ahead, induced by inaccurate speed estimates.
Although, our approach shows lower mean standard deviations, which verify the ability to predict velocities more uniformly.
\subsection{Driving Circles}
\begin{figure}
	\centering
	\setlength\fheight{3.8cm}
	\setlength\fwidth{7cm}
	\input{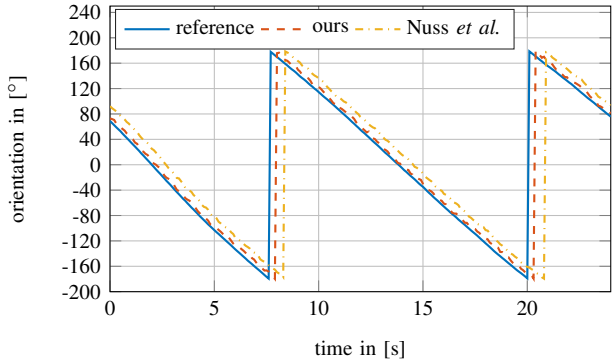}
	\caption[Driving circles velocity orientation]{Velocity orientation in the driving circles scenario.}
	\label{fig:DrivingCirclesOri}
\end{figure}
\begin{figure}
	\centering
	\begin{subfigure}{0.49\columnwidth}
		\includegraphics[width=\columnwidth, trim=0 140 0 140, clip]{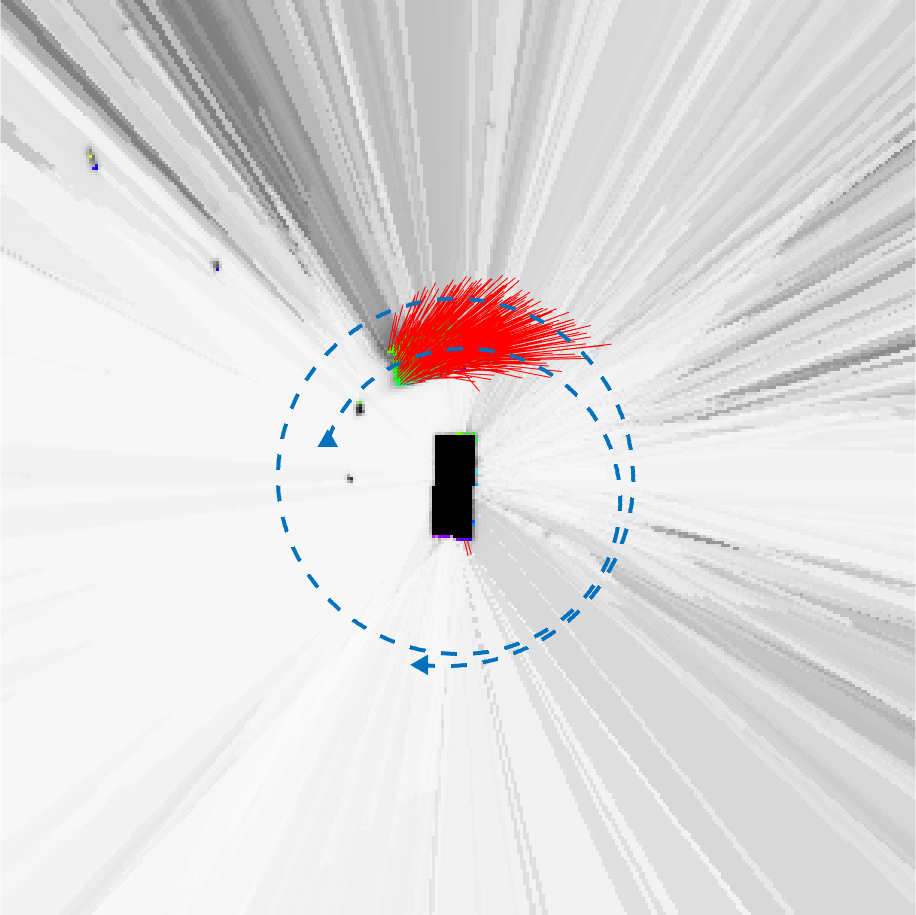}
		\caption{Nuss \textit{et~al.}}
	\end{subfigure}
	\begin{subfigure}{0.49\columnwidth}
		\includegraphics[width=\columnwidth, trim=0 140 0 140, clip]{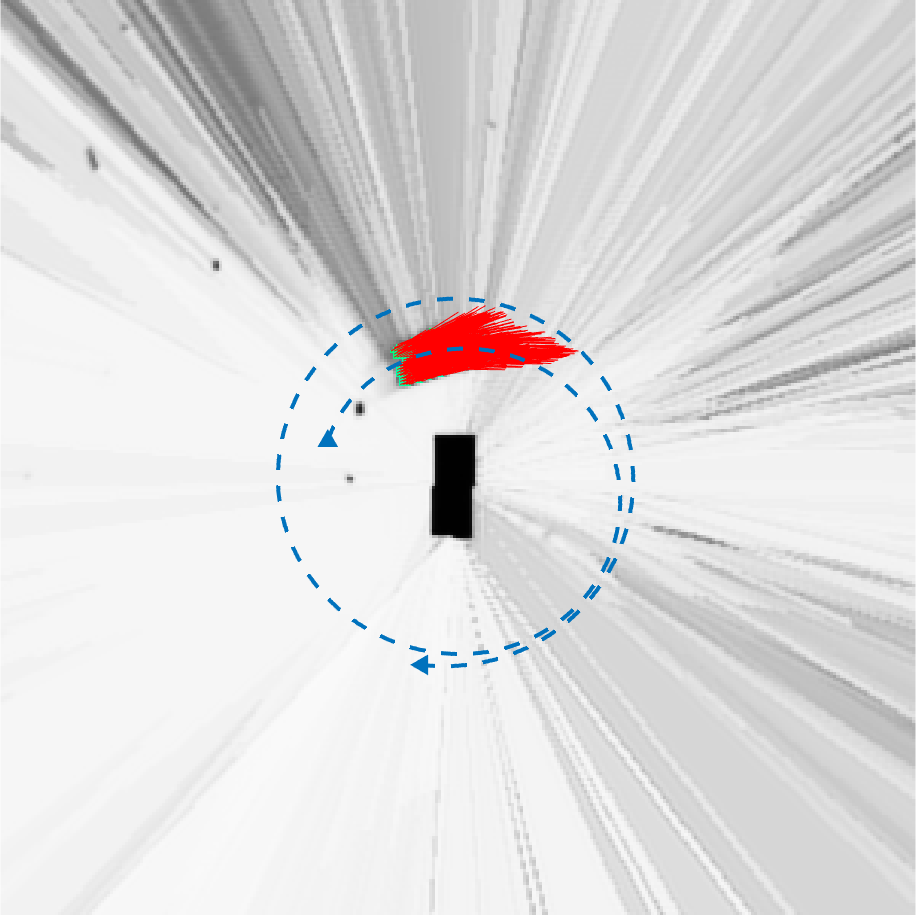}
		\caption{ours}
	\end{subfigure}
	\caption{Illustration of velocity estimates with red arrows drawn on top of the ground truth 
		     trajectory of the reference vehicle as blue dashed line.}
	\label{fig:DrivingCircleImage}	
\end{figure}
Here, we evaluate the velocity estimates in a scenario, where the reference vehicle drives circles around our ego-vehicle.
The scenario is illustrated in Fig.~\ref{fig:DrivingCircleImage} with the ground truth trajectory of the reference vehicle as blue line.
The orientation of the reference vehicle is depicted in blue in Fig.~\ref{fig:DrivingCirclesOri}.
The curves, as well as the mean orientation in Table~\ref{tab:EvalSummary} show, that our approach is able to estimate the continuous changing orientation much better than the particle filter.
In addition, $\bar{\sigma}\textsubscript{ori}$ is significant smaller, which is also recognizable in Fig. \ref{fig:DrivingCircleImage} with the velocities depicted as red arrows.
The left image shows that parts of the estimates of the particle filter approach point outward, while the network predictions point tangent to the curve.
Our approach outperforms the particle filter in this scenario in all metrics (see Table~\ref{tab:EvalSummary}).
\subsection{Qualitative Evaluation}\label{subsec:qualitativeEval}
\begin{figure}
	\vspace{1.42mm}
	\centering
	\begin{subfigure}{0.49\columnwidth}
		\includegraphics[width=\columnwidth]{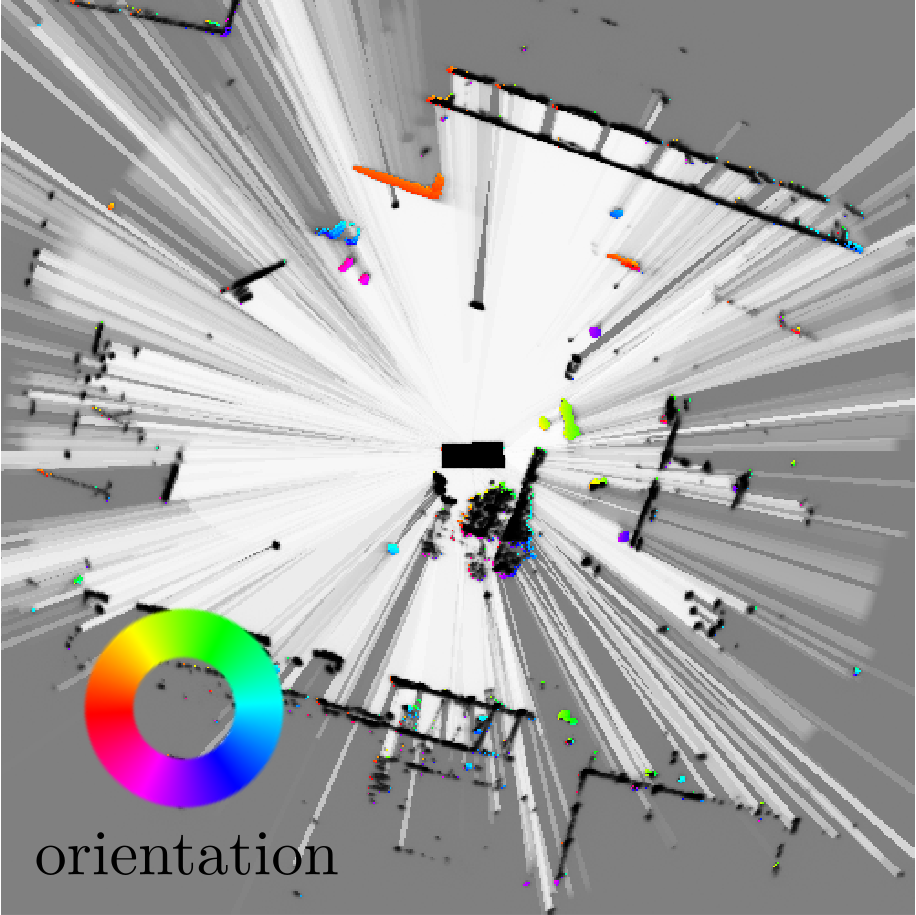}
	\end{subfigure}
	\begin{subfigure}{0.49\columnwidth}
		\includegraphics[width=\columnwidth]{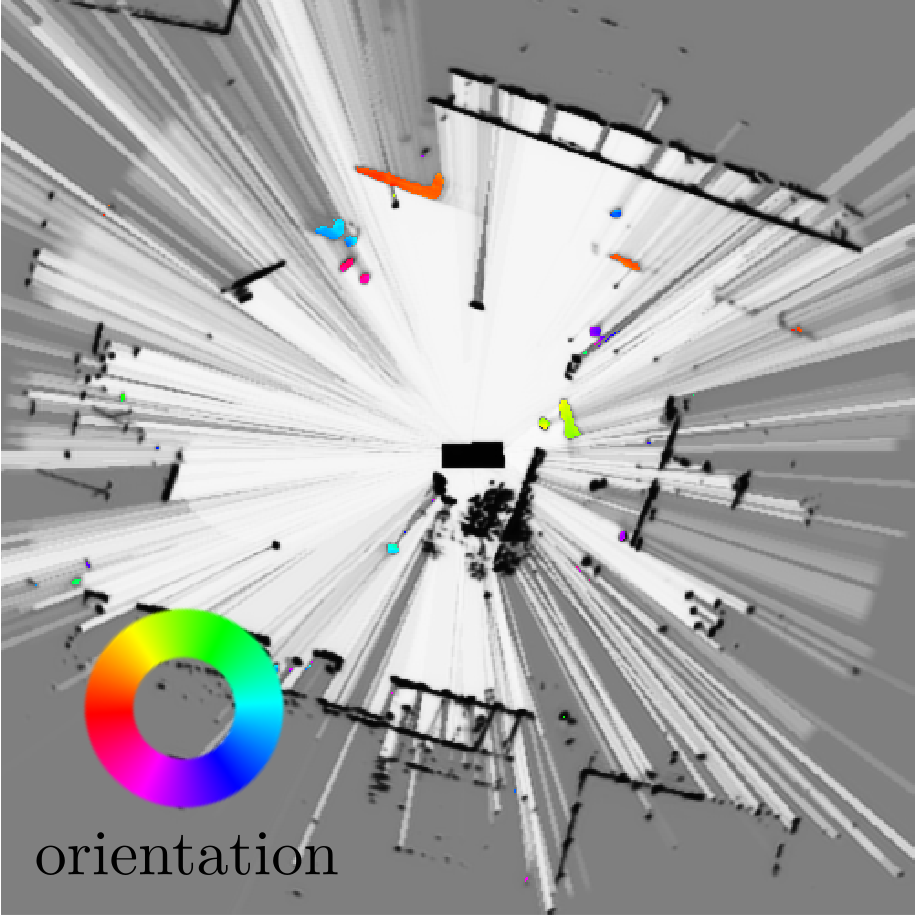}
	\end{subfigure}
	\begin{subfigure}{0.49\columnwidth}
		\vspace{0.9mm}
		\includegraphics[width=\columnwidth]{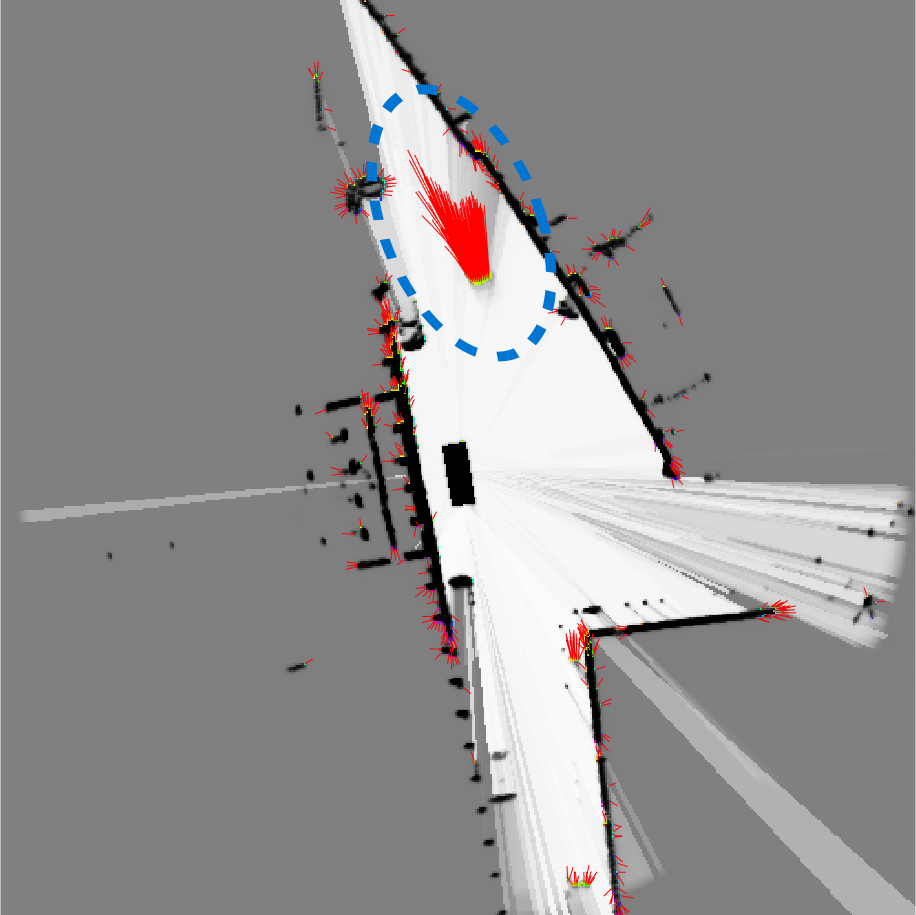}
		\caption{Nuss \textit{et~al.}}
	\end{subfigure}
	\begin{subfigure}{0.49\columnwidth}
		\vspace{0.9mm}
		\includegraphics[width=\columnwidth]{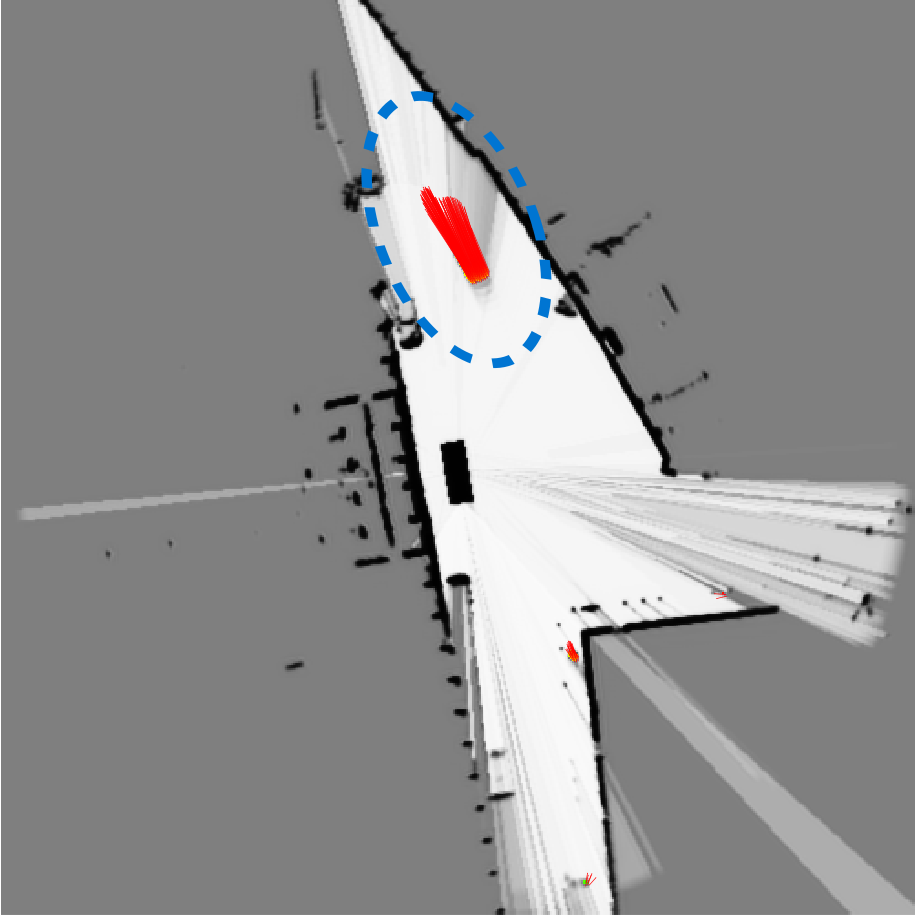}
		\caption{ours}
	\end{subfigure}
	\caption{Comparison of (a) particle filter based estimates and (b) our approach on the test set. In the illustration at the top, the orientation of the velocity estimates is depicted in colors according the colored circle, for velocities $>$ 0.7 m/s. In the bottom images the velocity estimates are visualized with red arrows for velocities $>$ 0.7 m/s.}
	\label{fig:qualitativeEval}	
\end{figure}
We provide a qualitatively comparison of our approach with the particle-filter approach \cite{DBLP:journals/corr/NussRTYKMGD16} from scenes of our test set.
Thereby, our test set comprises recordings of intersections, that differ from those of our train set (see Fig. \ref{fig:trainingdata}).
The two left images in Fig. \ref{fig:qualitativeEval} show dynamic occupancy grid maps predicted with (a) the algorithm in \cite{DBLP:journals/corr/NussRTYKMGD16} and (b) the predictions of our RNN.
In the top images, the velocity estimates are illustrated as colors according to their orientation.
This result demonstrates the strong generalization capability and robustness of our approach, since the predicted occupancy and velocity estimates of our RNN are very close to the ones of the particle filter approach.
Furthermore, it can be seen, that our approach produces less wrong velocity estimates in the static environment, i.e. there are less colors at contours of static obstacles.
In the two bottom grid maps in Fig.~\ref{fig:qualitativeEval}, the velocity estimates are visualized as red arrows. 
The arrows of the marked object show, that the orientation of the velocity estimates are much more uniform in the prediction of the RNN.
This result is also quantified in the evaluations above with the lower mean standard deviations in Table~\ref{tab:EvalSummary} of our approach.
The illustration also depicts, that the particle filter approach produces significant more false positives, i.e. predicting velocity in static area, than our approach.

In summary, our qualitative evaluations demonstrated the benefits of our training strategy, with the ability of our model to predict object motion more uniformly and produce less wrong velocity estimates in static area.
In addition, we compared our approach quantitatively with the algorithm in \cite{DBLP:journals/corr/NussRTYKMGD16} using three different driving scenarios.
The particle filter approach provided slightly better results in a setting with constant velocity and orientation.
However, in more complex driving scenarios with acceleration, brake and turn maneuvers our approach outperforms the particle filter.
\section{Conclusions}
In this work, we have presented a learning-based approach for predicting dynamic occupancy grid maps in stationary settings. 
Our recurrent neural network processes past measurement grid maps to predict the dynamic occupancy grid map at the current time step. 
The training of our network architecture takes place with automatically generated labels.
In the evaluation, we show that our approach is applicable in different urban environments.
Moreover, it provides promising results in dynamic driving situations, such as in brake and turn maneuvers, compared to the related work.
As future work, we aim to refine our approach for the application on a driving ego-vehicle.
%
%
%
%
\addtolength{\textheight}{-9.3cm}   



%
%
%
%
%
%
\bibliographystyle{IEEEtran}
\bibliography{IEEEtranControl,IEEEabrv,mybibfile}
\end{document}